\documentclass[journal]{IEEEtran}
\ifCLASSINFOpdf
\else
\fi
\usepackage{epsfig}
\usepackage{graphicx}
\usepackage{amsmath}
\usepackage{amssymb}
\usepackage{booktabs}
\usepackage{eso-pic}
\usepackage{graphicx} 

\begin{document}
%
\title{High-Order Paired-ASPP Networks for Semantic Segmentation}
%
%
%

\author{Yu~Zhang,
        Xin~Sun,~\IEEEmembership{Member,~IEEE,}
        Junyu~Dong,~\IEEEmembership{Member,~IEEE,}
        Changrui~Chen,
        Yue~Shen

\thanks{This work was supported in part by National Natural Science Foundation of China under Project No. 61971388, U1706218, L1824025. } 

\thanks{Y Zhang, X Sun, J Dong, C Chen and Y Shen are with the Department of Computer Science and Technology, Ocean University of China, Qingdao, Shandong Province, 266100 China (e-mail: sunxin1984@ieee.org)}

\thanks{Manuscript received April 19, 2005; revised August 26, 2015.}}

\maketitle

\begin{abstract}
Current semantic segmentation models only exploit first-order statistics, while rarely exploring high-order statistics. However, common first-order statistics are insufficient to support a solid unanimous representation. 
  In this paper, we propose High-Order Paired-ASPP Network to exploit high-order statistics from various feature levels. The network first introduces a High-Order Representation module to extract the contextual high-order information from all stages of the backbone. They can provide more semantic clues and discriminative information than the first-order ones. Besides, a Paired-ASPP module is proposed to embed high-order statistics of the early stages into the last stage. It can further preserve the boundary-related and spatial context in the low-level features for final prediction. Our experiments show that the high-order statistics significantly boost the performance on confusing objects. Our method achieves competitive performance without bells and whistles on three benchmarks, i.e, Cityscapes, ADE20K and Pascal-Context with the mIoU of 81.6\%, 45.3\% and 52.9\%.
\end{abstract}

\begin{IEEEkeywords}
Semantic Segmentation, Convolutional Neural Network, Scene Parsing, High-Order Statistics, Feature Representation.
\end{IEEEkeywords}

%
\IEEEpeerreviewmaketitle

\section{Introduction}
%
%
%
%

\IEEEPARstart{A}{s} a fundamental and challenging task in computer vision, semantic segmentation aims at predicting labels for every pixel accurately. FCNs \cite{Shelhamer2014FullyCN} based state-of-the-art approaches have achieved great success among several segmentation benchmarks. Currently, segmentation is widely applied to many vision-based applications such as autonomous driving \cite{Xu2016EndtoEndLO}, medical image analysis \cite{Litjens2017ASO} and remote sensing \cite{kampffmeyer2016semantic}. Nonetheless, these methods are inherently limited by the lack of discriminative information and the disparity between low-level and high-level features.

\begin{figure}[t]
\begin{center}
\includegraphics[width = 8cm, height = 5.5cm]{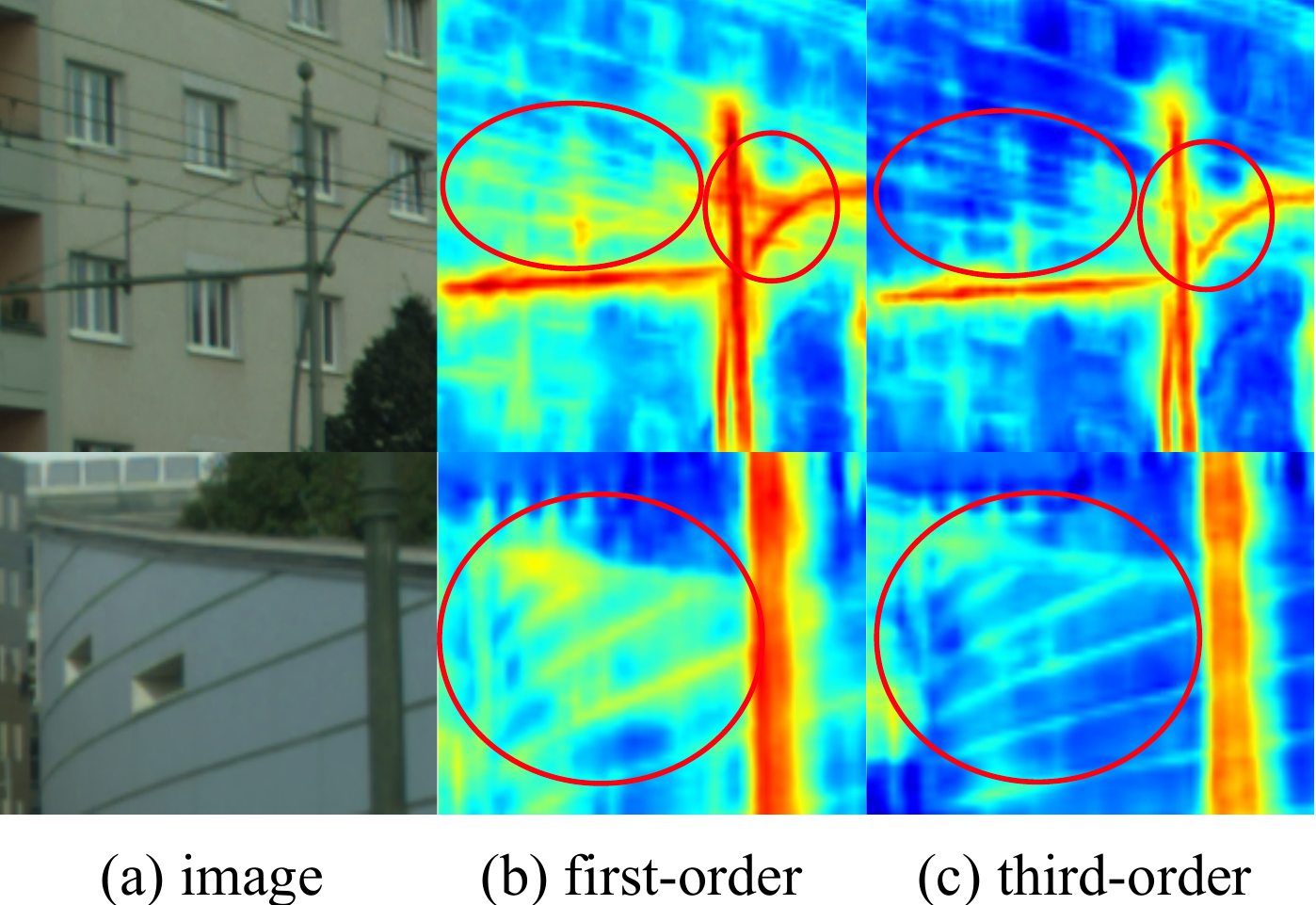}
\end{center}
   \caption{Illustration of the comparison between 1-$th$ and 3-$th$ order statistics. We can see that the confusing regions (in red circles) are corrected by high-order statistics. }
\label{introduction}
\end{figure}

Many state-of-the-art methods made great effort to address the above issues. Methods such as ASPP \cite{Chen2017RethinkingAC}, PSP \cite{Zhao2016PyramidSP}, Context Encoding \cite{Zhang2018ContextEF} and Self-Attention mechanism \cite{Vaswani2017AttentionIA, Fu2018DualAN, Huang2018CCNetCA,Li2019ExpectationMaximizationAN} are proposed to achieve global receptive fields and capture long-range context. They aggregate pixel-wise contextual information from high-level features to produce discriminative information. Besides, encoder-decoder structures \cite{zhang2018exfuse, li2019dfanet, takikawa2019gated, yu2018learning} are employed to retain and utilize spatial information from low-level features. They utilize feature fusion to reduce the disparity between low-level and high-level features. Although these methods improved the performance, they only focus on first-order information and high-level features. Their drawback is the neglect of high-order information and low-level features. The first-order information is coarse and insufficient to capture the semantic concepts of the confusing categories. For example, the 'electric wires' can be easily identified as 'pole', and the 'wall' represents similar features with the 'building'. However, high-order statistics can provide more discriminative information than first-order ones. As shown in Figure \ref{introduction}, the first-order statistics contain more noise than third-order ones, which may mislead the classifier. The third-order statistics remove some noise and reserve the edge information, improving the performance on the confusing categories.

Thus, in this paper we propose the {\bf High-Order Paired-ASPP Network} to exploit abstract statistical information from various level of layers. We first introduce a {\bf High-Order Representation} (HR) module to extract high-order statistics from different network layers. We adopt the polynomial kernel approximation based high-order methods \cite{Cai2017HigherOrderIO} to generate low-dimensional high-order statistics. The kernel representation can be reformulated with 1 $\times$ 1 convolution operation followed by element-wise product. With our HR module, the features will be mapped to a high-dimensional kernel feature space. The HR module can capture the differences between confusing categories and make features much more discriminative.

Furthermore, in order to utilize the low-level features, we propose a {\bf Paired-ASPP} (PA) module in the network to bridge the semantic and spatial information gap. The low-level features contain a large amount of boundary-related information, which contributes to dense pixel prediction. The PA module respectively combines the high-order statistics of the first three stages with the last one. The combination enables us to obtain multi-scale information and embed spatial resolution clues into high-level features, and each output contains two scales features. Then we apply 3 $\times$ 3 atrous convolution with various rates on these outputs to fuse different scales features. In this way we can effectively capture eight scales information.

In summary, the main contributions are:

\begin{itemize}
    \item [$\bullet$] We propose the High-Order Representation module to generate high-order statistics from different layers. To the best our knowledge, it is the first work to apply high-order information in semantic segmentation.
    \item [$\bullet$] We propose the Paired-ASPP module to fuse the high-order statistics of all levels. It takes full advantage of the low-level features and multi-scale receptive fields context.
    \item [$\bullet$] We integrate the HR and PA modules to construct the High-Order Paired-ASPP Network. The network enhances the richness of semantic clues and exploits information from the high-order statistics.
    \item[$\bullet$] We have conducted extensive experiments and achieved state-of-the-art performance on the Cityscapes, ADE20K and the PASCAL Context benchmarks. 
\end{itemize}
The rest of this paper is organized as follows. Related work is reviewed in Part II. And we present the proposed High-order Representation module and Paired-ASPP module in Part III. Experimental results are presented in Part IV. Finally, Part V concludes this paper.

\section{Related Works}

In recent years, the developments of deep neural networks encourage the emergence of a series of works on semantic segmentation. FCN \cite{Shelhamer2014FullyCN} is the first approach to adopt fully convolution network for semantic segmentation. Later, state-of-the-art segmentation approaches based on FCN \cite{Shelhamer2014FullyCN} have made great progress in semantic segmentation.

\subsection{Encoder-Decoder}
A encoder generally reduces the spatial size of feature maps to get semantic context information and enlarge the receptive field. Then the decoder recovers the predictions. As the first encoder-decoder approach on semantic segmentation, FCN \cite{Shelhamer2014FullyCN} shows good performance. It discards the fully connected layer to support input of arbitrary sizes. Based on FCN \cite{Shelhamer2014FullyCN}, many works change the upsample methods to generate better results.  The SegNet \cite{badrinarayanan2017segnet} stores the max-pooling indices of the feature maps and uses them in its decoder network to make upsample convincing. The DeconvNet \cite{noh2015learning} uses deconvolutions to perform the decoding pass. Besides, multi-level feature fusion in semantic segmentation is a hot topic, many encoder-decoder methods try to solve this problem. The U-Net \cite{ronneberger2015u} gradually fuses high-level low-resolution features with low-level but high-resolution features, which is helpful for decoder to generate better results. This architecture is adopted in many works \cite{ghiasi2016laplacian, amirul2017gated, lin2017refinenet}. RefineNet \cite{lin2017refinenet} exploits all the information available along the down-sampling process to enable high-resolution predictions using long-range residual connections. ICNet \cite{zhao2018icnet} incorporates multi-resolution branches under proper label guidance to address the challenge of real-time semantic segmentation. FCDenseNet \cite{jegou2017one} adjust DenseNet \cite{Huang2016DenselyCC} for semantic segmentation. ExFuse \cite{zhang2018exfuse} assigns auxiliary supervisions to all stages to force low-level features to encode more semantic concepts and high-level features to learn more spatial information. DFN \cite{yu2018learning} uses two sub-networks, one smooth network and one border network to select the more discriminative features for later segmentation. Besides, Conditional Random Field \cite{lafferty2001conditional} can be used in encoder-decoder structures for semantic segmentation. Combining with CNNs, the methods \cite{chandra2016fast, Chen2014SemanticIS, he2015spatial, chen2017deeplab, vemulapalli2016gaussian} adopted this strategy, making the deep network end-to-end trainable. The encoder-decoder structure guarantees the competitive performance. However, it is difficult to discover disciminative feature and fuse the different level features using only simple encoder-decoder structure.

\subsection{Context Embedding}
Context embedding has shown promising improvements on semantic segmentation. GCN \cite{Peng2017LargeKM} utilizes $1 \times k$, $k \times 1$ convolution to constitude a global convolution block. ParseNet \cite{liu2015parsenet} uses global pooling to clarify local confusions. Both of these two methods are intended to enlarge the receptive field and capture context information.  Recently, self-attention mechanism \cite{Vaswani2017AttentionIA} is widely used for semantic segmentation to get global dependencies. The self-attention \cite{Vaswani2017AttentionIA} is originally proposed to solve the machine translation \cite{shen2018disan}, \cite{lin2017structured}, and the following work \cite{wang2018non} further proposed the non-local neural network for various tasks such as object detection, video classification and instance segmentation. DANet \cite{Fu2018DualAN} appends two types of attention modules which model the semantic interdependencies in spatial and channel dimensions. CCNet \cite{Huang2018CCNetCA} harvest the contextual information of its surrounding pixels on the criss-cross path through a novel criss-cross attention module. As variants of self-attention structure, EMANet \cite{Li2019ExpectationMaximizationAN} and ANN \cite{zhu2019asymmetric} present new self-attention modules to reduce the computation and memory consumption without sacrificing the performance. To extract the multi-scale information and global context efficiently, SPP \cite{he2015spatial} based methods are proposed. Deeplabv2 \cite{chen2017deeplab} proposes ASPP module, where parallel atrous convolution layers with different rates to capture multi-scale information, then concatenate the different scales features to fuse the local and global context. PSPNet \cite{Zhao2016PyramidSP} performs spatial pooling at several grid scales to aggregate global contextual information, the local and global clues together make the final prediction more reliable. The ASPP module is improved in DeepLabV3 \cite{Chen2017RethinkingAC} by integrating a global pooling branch. DenseASPP \cite{yang2018denseaspp} connects a set of atrous convolutional layers in a dense way, such that it generates multi-scale features that not only cover a larger scale range, but also cover that scale range densely, besides the module doesn't increase the computation cost too much. All of the SPP \cite{he2015spatial} variants are stacked at the top of their backbone networks for prediction, which ignores the spatial information. These works achieve great performance on semantic segmentation.  However, the low-level and high-level features in these methods are first-order and lack of spatial information. Thus, we propose the High-Order Representation module and Paired-ASPP module to exploit high-order statistics and capture more discriminative spatial information. \\

\subsection{High-order statistics}
Recently, some works of image classification and recognition \cite{cai2017higher, cui2017kernel, ionescu2015matrix, lin2015bilinear, li2017second} show that the integration of high-order statistics significantly improves their performance. Global high-order pooling is utilized to aggregate the information and represent whole images \cite{ionescu2015matrix, lin2015bilinear, li2017second}, in which the sum of outer product of convolutional features is firstly computed, then element-wise power normalization \cite{lin2015bilinear}, matrix logarithm normalization \cite{ionescu2015matrix} and matrix power normalization \cite{li2017second} are performed. G2DeNet \cite{Wang2017G2DeNetGG} adds a trainable layer of one global Gaussian as an image representation into CNNs, which exploits first-order and second-order information. All these methods generate very high dimensional representations, which is not suitable for other tasks such as semantic segmentation. Some works \cite{cai2017higher, cui2017kernel} adopt polynomial and Gaussian RBF Kernel functions to generate low-dimensional high-order representations. MLKP \cite{wang2018multi} proposes a location-aware kernel approximation method to exploits high-order statistics for improving performance of object detection. MHN \cite{chen2019mixed} proposes the High-Order Attention module, using complex and high-order statistics to increase the discrimination of proposals. It improves the performance of person-ReID task. All these methods only extract high-order information from single scale. In this paper, we generate high-order features from different layers and fuse them. Therefore, our high-order statistics contain plenty of spatial details and semantic concepts. To the best of our knowledge, our method is the first work to apply high-order information in semantic segmentation.

\section{High-Order Paired-ASPP Network}
In this section, we introduce the High-Order Paired-ASPP Network. Sec ~\ref{HRModule} firstly introduces the High-Order Representation module. Then Sec \ref{HFFN} shows the Paired-ASPP module. Finally we illustrate the overall framework of our High-Order Paired-ASPP Network in Sec \ref{NA}.
\subsection{High-Order Representation Module}\label{HRModule}
To get high-order representations from each stage, we first let $\mathcal{X} \in \mathbb{R}^{C \times M \times N } $ be the feature map extracted from the given convolutional layer, where $C$, $H$ and $W$ indicate the number of channel, height and width. Then we can define a linear predictor on the high-order statistics of $\mathrm{x}$, where $\mathrm{x} \in \mathbb{R}^C $ denotes a local descriptor from $\mathcal{X}$.
\begin{equation}\label{equation1}
    f(\mathrm{x}) = \langle \mathrm{w}, \phi(\mathrm{x})\rangle,
\end{equation}
\noindent
where $\mathrm{w}$ is the parameter of predictor, $\phi(\mathrm{x})$ denotes the high-order statistics. We can reformulate Eq \ref{equation1} with a homogeneous polynomial kernel as
\begin{equation}\label{equation2}
    f(\mathrm{x}) = \sum_{r=1}^{R} \langle \mathrm{w^{r}}, \otimes_r \mathrm{x}\rangle.
\end{equation}
The $\langle\cdot, \cdot\rangle$ is inner product of two same-sized tensors. $R$ is the number of order, tensor $\otimes_r \mathrm{x}$ comprises all the degree-$r$ monomials in $\mathrm{x}$, and $r$-th order tensor $\mathrm{w^{r}}$ determines a degree-$r$ homogenous polynomial predictor.

\begin{figure}[t]
\begin{center}
\includegraphics[width=0.45\textwidth]{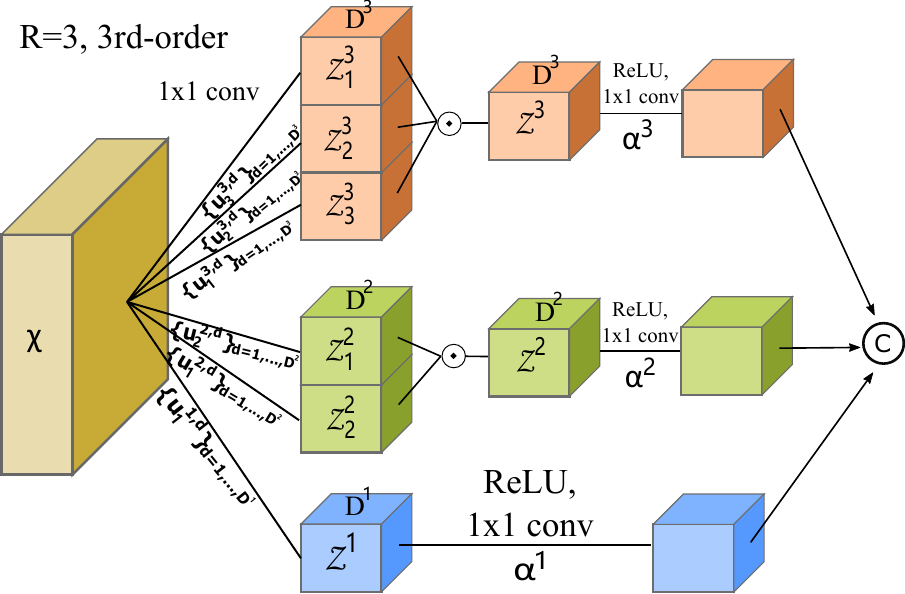}
\end{center}
   \caption{Illustration of the \textbf{High-Order Representation} module. This is the example of order $R = 3$. $\mathcal{X}$ is the input feature map. We apply six $1 \times 1$ convolution operations followed by element-wise product on $\mathcal{X}$. Then  $\mathcal{Z}^1,\mathcal{Z}^2, \mathcal{Z}^3$ are implemented by $1 \times 1$ convolution with ReLU activation. Finally we concatenate three feature maps to generate third-order representation.}
\label{hrmodule}
\end{figure}

With the tensor decomposition \cite{kolda2009tensor}, $\mathrm{w^{r}}$ can be approximated by $D^r$ rank-1 tensors, i.e., $\mathrm{w^{r}} = \sum_{d=1}^{D^r}\alpha^{r,d}u_{1}^{r,d}\otimes \cdots \otimes u_{r}^{r,d} $ in case of $r > 1$, where $a^{r,d}$ is the weight for $d$-th rank-1 tensor, $u_{1}^{r,d}, \cdots,  u_{r}^{r,d} \in \mathbb{R}^C$ are vectors, and $\otimes$ is the outer-product. The Eq \ref{equation1} can be further rewritten as follow.
\begin{equation}\label{equation3}
\begin{split}
    f(\mathrm{x}) &=\langle \mathrm{w^{1}}, \mathrm{x} \rangle + \sum_{r=2}^{R} \langle \sum_{d=1}^{D^r}\alpha^{r,d}u_{1}^{r,d}\otimes \cdots \otimes u_{r}^{r,d}, \otimes_r  \mathrm{x}  \rangle \\
    &= \langle \mathrm{w^{1}}, \mathrm{x} \rangle + \sum_{r=2}^{R} \sum_{d=1}^{D^r} \alpha^{r,d} \prod_{s=1}^{r} \langle u_{s}^{r,d}, \mathrm{x} \rangle \\
    &= \langle \mathrm{w^{1}}, \mathrm{x} \rangle + \sum_{r=2}^{R} \langle \alpha^r, \mathrm{z}^r \rangle,
\end{split}
\end{equation}
\noindent
where $\alpha^r$ is the weight vector and $\mathrm{z}^r = [z^{r,1}, \cdots, z^{r, D^r}]^T$ with $z^{r,d} = \prod_{s=1}^{r}\langle u_{s}^{r,d}, \mathrm{x} \rangle$.

With the Eq \ref{equation3}, we can calculate any arbitrary order of representation by learning the parameters of weight $\mathrm{w^{1}}$, $\alpha^r$ and $u_{s}^{r,d}$. For the $z^{r,d} = \prod_{s=1}^{r}\langle u_{s}^{r,d}, \mathrm{x} \rangle$, we can first deploy $\{u_{s}^{r,d}\}_{d=1, \cdots, D_r}$ as a set of $D_r$ $1 \times 1$ convolutional filters on $\mathcal{X}$ to generate a set of feature maps $\mathcal{Z}_{s}^r$ of dimension $D^r\times  M\times N$. In general, the dimension of $D^r$ should be large enough to make a good approximation. However, in practice, we should use a relative low dimension because the high-dimension increases the computational cost and causes the over-fitting problem. The feature maps $\{ \mathcal{Z}_{s}^{r} \}_{s=1, \cdots, r}$ are combined by element-wise product to get $\mathcal{Z}^r = \mathcal{Z}_{1}^{r} \odot \cdots \odot \mathcal{Z}_{r}^{r} $. The $\{u_{s}^{r,d}\}_{d=1, \cdots, D_r}$ can be regarded as a polynomial module to learn degree-r polynomial features.

\begin{figure*}[t]
\centering
\includegraphics[width = 17cm, height = 6.5cm]{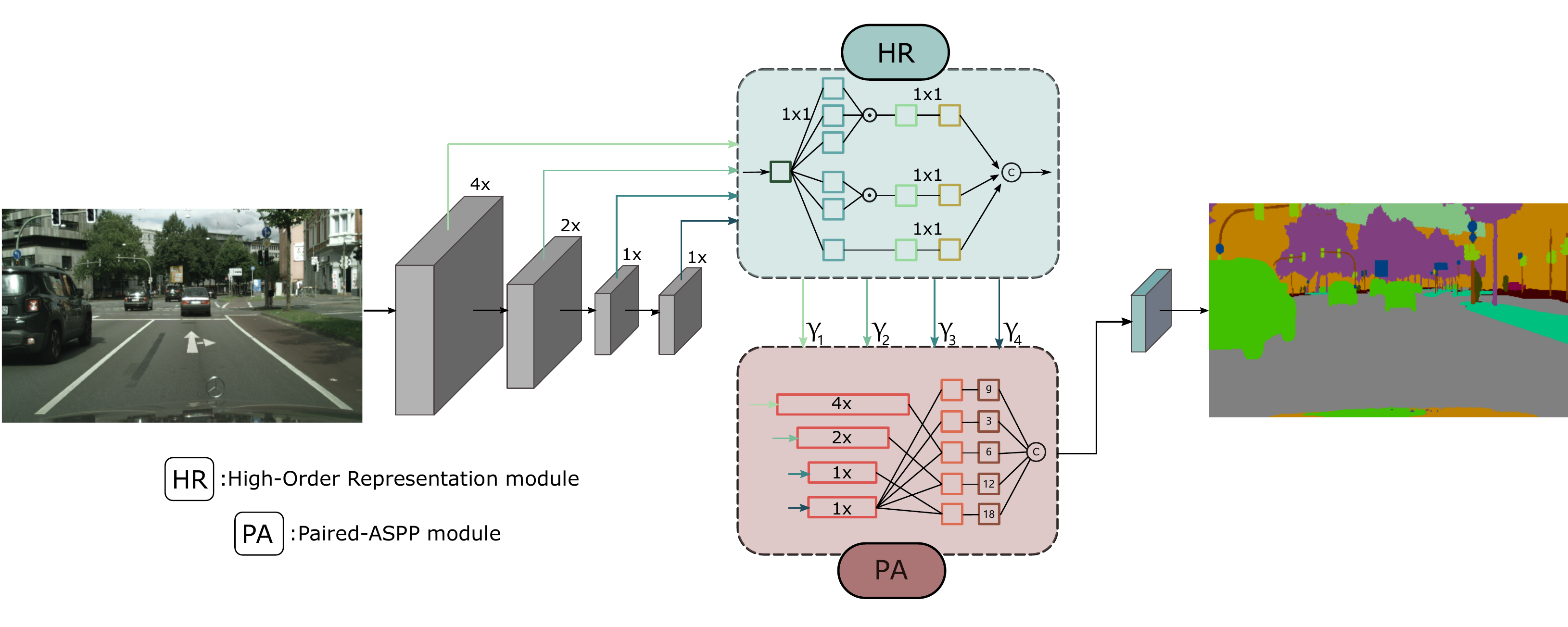}
   \caption{The architecture of our \textbf{High-Order Paired-ASPP Network}. Each of stages in backbone generates high-order statistics $\mathcal{Y}_1, \mathcal{Y}_2, \mathcal{Y}_3, \mathcal{Y}_4$ through four HR modules. Then the high-order statistics are fused by one PA module.}
\label{architecture}
\end{figure*}

\begin{figure}[t]
\begin{center}
\includegraphics[width=0.41\textwidth]{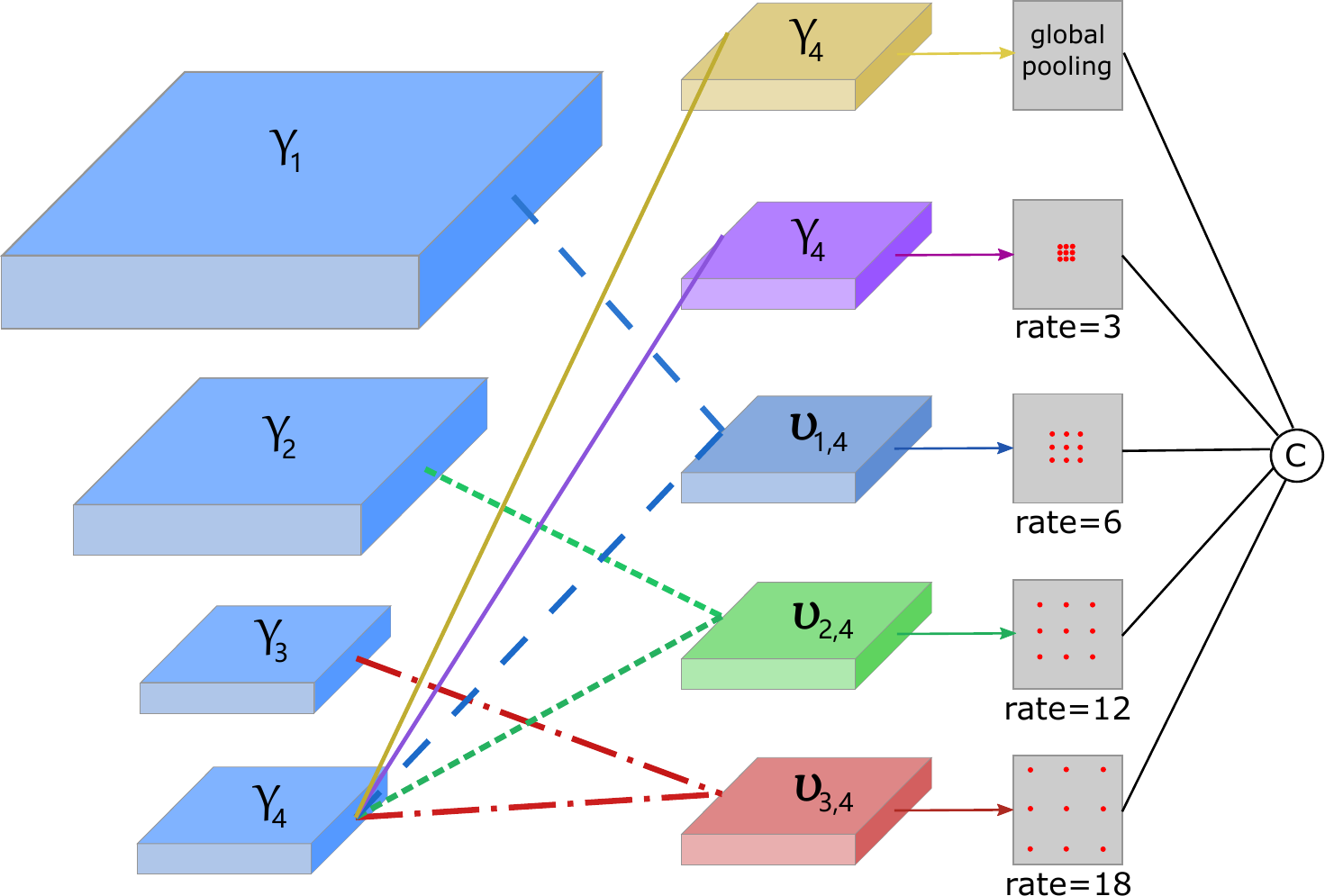}
\end{center}
   \caption{Illustration of \textbf{Paired-ASPP} module. $\mathcal{Y}_1$, $\mathcal{Y}_2$, $\mathcal{Y}_3$, and $\mathcal{Y}_4$ are four outputs from each stage of the backbone. To make structure clearer, we use different colored lines to represent different connections.  }
\label{Paired-ASPP}
\end{figure}

\textbf{Implementation}:\quad As shown in Figure \ref{hrmodule}, the HR module of $R = 3$ can be easily implemented. At each stage of the backbone, we can get a feature map $\mathcal{X}$. Then we adopt $\{u_{s}^{r,d}\}_{d=1, \cdots, D_r}$ as a set of $1 \times 1$ convolution layers on $\mathcal{X}$. So we can get $6$ representations with channels $D^r$, $\mathcal{Z}_{1}^{1}, \cdots, \mathcal{Z}_{3}^{3}$. These feature maps are combined by element-wise product to generate $\mathcal{Z}^1,\mathcal{Z}^2$, and $\mathcal{Z}^3$. And $\alpha^1, \alpha^2, \alpha^3$ can also be implemented by three $1 \times 1$ convolution layers. Finally we concatenate the three feature maps as the third-order statistics of this stage.

\subsection{Paired-ASPP Module}\label{HFFN}
In a general encoder-decoder structure, high-level features of the encoder output depict the semantic information of the input images. Although the highest-level features contain much semantic information, it is also necessary to make use of the low-level features. Neglecting low-level features leads to the lack of multi-scale information and concrete spatial details. To extract discriminative information from features in shallow layers, we introduce the Paired-ASPP (PA) module. The PA module collects high-order statistics from all stages of the backbone and fuses them via a pairing manner.

The normal ASPP module applies four parallel atrous convolutions with different atrous rates on the last stage. This enable the ASPP module to extract multi-scale information and whole image context. We argue that only using the highest-level feature is inappropriate. Therefore, we combine the high-order statistics of the first three stages to the last stage in a pairing manner. In this way we can capture more scales information than normal ASPP while retaining the spatial information from low-level layers.

As shown in Figure \ref{Paired-ASPP}, we get four feature maps $\mathcal{Y}_1$, $\mathcal{Y}_2$, $\mathcal{Y}_3$, and $\mathcal{Y}_4$ where $\{\mathcal{Y}_s\}_{s=1, \cdots, 4}$ is the high-order representation of each stage. Then we concat the first three feature maps $\mathcal{Y}_1, \mathcal{Y}_2, \mathcal{Y}_3$ with $\mathcal{Y}_4$ in a pairing manner:
\begin{equation}\label{equation4}
\begin{split}
    &\mathcal{V}_{1,4} = Concat(\mathcal{Y}_1, \mathcal{Y}_4), \\
    &\mathcal{V}_{2,4} = Concat(\mathcal{Y}_2, \mathcal{Y}_4), \\
    &\mathcal{V}_{3,4} = Concat(\mathcal{Y}_3, \mathcal{Y}_4).
\end{split}
\end{equation}

We can regard the $\mathcal{V}_{1,4}, \mathcal{V}_{2,4}, \mathcal{V}_{3,4}$ as multi-view information of the scene, each of which contains two scales high-order statistics. We apply four atrous $3 \times 3$ convolutions with $rates = (3, 6, 12, 18)$ and one global pooling on $\mathcal{V}_{1,4}, \mathcal{V}_{2,4}, \mathcal{V}_{3,4}$ and $\mathcal{Y}_4$. In this way, each atrous convolution can capture information of two scales. We want to use different atrous rates convolutions to cover the span of scales as large as possible. Therefore, we apply atrous convolution with the biggest rate 18 on $\mathcal{V}_{3,4}$ since  the receptive field of $\mathcal{Y}_3$ and $\mathcal{Y}_4$ is large. Meanwhile, atrous convolution with the small rates ${6, 12}$ is employed on $\mathcal{V}_{1,4}, \mathcal{V}_{2,4}$. The main reason is that low-level features need low-rate atrous convolutions to maintain their spatial information. Moreover, we apply normal convolution and global average pooling on original $\mathcal{Y}_4$. The normal convolution also extracts one scale information and the global average pooling captures global context information.

\begin{equation}\label{equation5}
\begin{split}
    &\mathcal{U}_{1} = Conv_{rates=18}(\mathcal{V}_{3,4}), \\
    &\mathcal{U}_{2} = Conv_{rates=12}(\mathcal{V}_{2,4}), \\
    &\mathcal{U}_{3} = Conv_{rates=6}(\mathcal{V}_{1,4}), \\
    &\mathcal{U}_{4} = Conv_{rates=1}(\mathcal{Y}_{4}), \\
    &\mathcal{U}_{5} = GAP(\mathcal{Y}_{4}). 
\end{split}
\end{equation}

Then we concatenate $\{\mathcal{U}_i\}_{i=1,\cdots,5}$. It is followed by several convolutional layers with batch normalization and ReLU activation. Finally the fused features are fed into segmentation layer to give the prediction.

\subsection{Network Architecture} \label{NA}
The architecture of our High-Order Paired-ASPP Network is shown in Figure \ref{architecture}. We choose the ResNet \cite{he2016deep} as our backbone network. We remove the last two stages with atrous convolution and make both strides to 1 in order to keep the spatial information. Each stage of the ResNet has an output $\mathcal{X}_i$.  The four feature maps $\{\mathcal{X}_i\}_{i=1,2,3,4}$ are used as the input for four High-Order Representation modules (R=3) to capture high-order statistics $\{\mathcal{Y}_i\}_{i=1,2,3,4}$. Then, all the high-order statistics $\{\mathcal{Y}_i\}_{i=1,2,3,4}$ are sent to the Paired-ASPP module for multi-scale feature extraction and feature fusion. All the convolutional layers are followed with batch normalization and ReLU activation. In the Paired-ASPP module, all the feature maps are concatenated finally. And we will use it for final semantic segmentation. Comparing with other methods which use segmentation heads, our proposed method models the high-order disparity between confusing objects. Besides we take full advantage of low-level features, making high-level features more comprehensive and discriminative. The entire network is trained in an end-to-end manner driving by cross-entropy loss defined on the segmentation benchmarks.

\section{Experiments}
In this section, we evaluate the proposed network by carrying out extensive experiments on the Cityscapes, ADE20K and the PASCAL Context datasets. Results show that the High-Order Paired-ASPP network achieves state-of-the-art performance. In following subsections we will first introduce the datasets and implementation details. Then we show the comparison results on the three datasets. At last, we carry out the ablation experiments.

\subsection{Datasets and Evaluation Metrics}

\noindent
\textbf{Cityscapes} \cite{cordts2016cityscapes} is a large-scale dataset to train and test approaches for semantic and instance segmentation, which contains 5000 high quality pixel-level annotations and 20000 coarse annotations. All the images come from a set of video sequences recorded in the streets from 50 different cities and the image size is $1024 \times 2048$ resolution. The fine data contains 2975 training, 500 validation and 1525 test images. And there are 30 classes in the dataset but only 19 classes are used for evaluation. \\

\noindent
\textbf{ADE20K} \cite{zhou2017scene} is a challenging scene parsing benchmark covering a wide range of scenes and object categories with dense and detailed annotations. The dataset contains 150 classes and all images are divided into 20K/2K/3K for training, validation and testing. Besides the dataset parses the scene into stuff, objects, and object parts, bringing more difficulties for semantic segmentation methods. \\

\noindent
\textbf{PASCAL Context} \cite{mottaghi2014role} labels every pixel of PASCAL VOC 2010 detection challenge with a semantic category, containing 4998 images for training and 5105 images for validation. We use 60 categories (59 object categories and background) for evaluation. \\

\noindent
\textbf{Evaluation Metric.} Our evaluation metric is mean of class-wise intersection over union (mIoU).

\subsection{Implementation Details}

\noindent
\textbf{Network Structure.} Our baseline network is ResNet-based FCN with dilated convolution module incorporated at stage 4 and 5.  Following \cite{Chen2017RethinkingAC}, we remove the last two down-sample operation in the ResNet, so that the last stage feature map is $8\times$ smaller than the input image. To get the final semantic labels for each pixel, we use bilinear interpolation to upsample the feature map to the target size. Besides, we use our own PyTorch implementation of DeepLabV3 \cite{Chen2017RethinkingAC}. \\

\noindent
\textbf{Training Settings.} We use the PyTorch framework and replace normal batch normalization with cross-GPU synchronize batch normalization for better mean/variance estimation. We adopt Stochastic Gradient Descent (SGD) with warm-up poly learning rate policy to optimize our network. Different initial learning rates, weight decays, batch sizes, crop sizes and training iterations are applied to different datasets. For Cityscapes and PASCAL Context, we train our network for 60K/28K iterations with the initial learning rate 0.01. And we choose the momentum of 0.9 and a weight decay of 0.0005, the batch size of 8/16 and the crop size of $864 \times 864$/ $480 \times 480$. For ADE20K, we train our network for 150K iterations with the initial learning rate of 0.02, the momentum of 0.9 and a weight decay of 0.001, the batch size is 16 and crop size is $512 \times 512$. For all the datasets, we adopt random scaling in the range of [0.5, 2.0], random horizontal flip and random brightness as additional data augmentation methods. All the experiments are conducted on $4\times$ RTX 2080Ti GPUs. \\

\noindent
\textbf{Inference Settings}
For inference, following other state-of-the-art methods, we apply multi-scale and horizontal flip testing for the Cityscapes, ADE20K and the PASCAL Context. The scales is in the range of [0.5, 1.75] with $stride = 0.25$.

\begin{table}[h]
\begin{center}

\caption{Comparison results on Cityscapes validation set. }
\begin{tabular}{c|c|c|c}
\toprule
\midrule
Method & Backbone & MS\_F & mIoU(\%) \\
\hline
DeepLabV3+ \cite{chen2018encoder} & Xception-65 &  & 79.1\\
DFN \cite{yu2018learning} & ResNet-101 & & 79.9\\
ANN \cite{zhu2019asymmetric} & ResNet-101 & & 80.1\\
DPC \cite{chen2018searching}$\dagger$ & Xception-71 &  & 80.8 \\
\hline
Ours & ResNet-101 & & 80.7 \\
\hline
\hline
DeepLabV3 \cite{Chen2017RethinkingAC} & ResNet-101 & $\surd$ & 79.3\\
DFN \cite{yu2018learning} & ResNet-101 & $\surd$ & 80.1\\
CCNet \cite{Huang2018CCNetCA} & ResNet-101 & $\surd$ & 81.3\\
\hline
Ours & ResNet-101 & $\surd$ & \textbf{81.8}\\
\midrule
\bottomrule
\end{tabular}
\end{center}

\label{cityscapesval}
\end{table}

\subsection{Comparison with State-of-the-art Methods}

\subsubsection{Cityscapes Dataset} 
We first illustrate the performance of our method on Cityscapes dataset. We directly train our High-Order Paired-ASPP Network for 60K iterations \textbf{without coarse data}. The comparison results of our method and the state-of-the-art methods on Cityscapes validation set are summarized in Table \ref{cityscapesval}.  

Among these approaches, DeepLabV3+ and DPC use more stronger backbone than ours, and DPC utilizes COCO dataset for training. As shown in Table \ref{cityscapesval}, the results show that our network outperforms others in validation set.

In addition, we also train our network with both training-fine and val-fine sets, and evaluate on test set by \textbf{submitting the results to the official evaluation server}. Most of the methods use the same backbone ResNet-101 as ours. ResNet-38 and DenseASPP \cite{wu2019wider, yang2018denseaspp} use stronger backbones than others. Besides DeepLabV3 \cite{Chen2017RethinkingAC} uses both fine and coarse sets for training. As shown in Table \ref{cityscapestest}, our network achieves $81.5\%$ mIoU, which outperforms the others. DeepLab \cite{chen2017deeplab, Chen2017RethinkingAC} and PSPNet \cite{Zhao2016PyramidSP} adopt multi-scale context fusion module to aggregate contextual information. CCNet \cite{Huang2018CCNetCA} and ANN \cite{zhu2019asymmetric} utilize self-attention mechanism to obtain long-range dependencies. These methods also achieve great performance on the Cityscapes dataset. However, these methods only utilize the first-order information. In contrast, our High-Order Paired-ASPP Network adopts high-order statistics for semantic segmentation and achieves the highest performance.

\begin{table}[ht]
\begin{center}

\caption{Comparison results on Cityscapes test set. $\dagger$ train with training-fine/val-fine datasets. $\ddagger$ train with fine/coarse datasets.}
\begin{tabular}{c|c|c}

\toprule
\midrule
Method & Backbone & mIoU(\%) \\
\hline
DeepLabV2 \cite{chen2017deeplab} & ResNet-101 & 70.4\\
RefineNet \cite{lin2017refinenet}$\dagger$ & ResNet-101 & 73.6 \\
GCN \cite{Peng2017LargeKM}$\dagger$ & ResNet-101 & 76.9 \\
DUC \cite{wang2018understanding}$\dagger$ & ResNet-101 & 77.6 \\
SAC \cite{zhang2017scale}$\dagger$ & ResNet-101 & 78.1 \\
ResNet-38 \cite{wu2019wider}$\dagger$ & WiderResNet-38 & 78.4 \\
PSPNet \cite{Zhao2016PyramidSP}$\dagger$ & ResNet-101 & 78.4 \\
BiSeNet \cite{yu2018bisenet}$\dagger$ & ResNet-101 & 78.9 \\
AAF \cite{ke2018adaptive}$\dagger$ & ResNet-101 & 79.1 \\
DFN \cite{yu2018learning}$\dagger$ & ResNet-101 & 79.3 \\
PSANet \cite{zhao2018psanet}$\dagger$ & ResNet-101 & 80.1 \\
DenseASPP \cite{yang2018denseaspp}$\dagger$ & DenseNet-161 & 80.6 \\
PSPNet \cite{Zhao2016PyramidSP}$\ddagger$ & ResNet-101 & 81.2 \\
ANN \cite{zhu2019asymmetric}$\dagger$ & ResNet-101 & 81.3 \\
DeepLabV3 \cite{Chen2017RethinkingAC}$\ddagger$ & ResNet-101 & 81.3 \\
CCNet \cite{Huang2018CCNetCA}$\dagger$ & ResNet-101 & 81.4 \\
\hline
Ours$\dagger$ & ResNet-101 & \textbf{81.6}\\
\midrule
\bottomrule
\end{tabular}
\end{center}

\label{cityscapestest}
\end{table}

\subsubsection{ADE20K dataset}
We conduct experiments on the ADE20K dataset to further prove the effectiveness of our proposed network. We compare our network to the previous state-of-the-art methods in Table \ref{ADE20Kval}. We can see that our method boosts the performance to $45.30\%$. EncNet \cite{Zhang2018ContextEF} achieves good performance among the methods by using global pooling and image-level supervision to capture the global semantic context of scenes. However, our High-Order Paired-ASPP Network provides a new way of obtaining multi-scale contextual information and achieve better performance than EncNet \cite{Zhang2018ContextEF}.

\begin{table}[h]
\begin{center}

\caption{Comparison results on ADE20K validation set.}
\begin{tabular}{c|c|c}

\toprule
\midrule
Method & Backbone & mIoU(\%) \\
\hline
RefineNet \cite{lin2017refinenet} & ResNet-152 & 40.70\\
UperNet \cite{xiao2018unified} & WiderResNet-38 & 42.66 \\
PSPNet \cite{Zhao2016PyramidSP} & ResNet-101 & 43.29 \\
DSSPN \cite{liang2018dynamic} & ResNet-101 & 43.68 \\
PSANet \cite{zhao2018psanet} & ResNet-101 & 43.77 \\
SAC \cite{zhang2017scale} & ResNet-101 & 44.30 \\
EncNet \cite{Zhang2018ContextEF} & ResNet-101 & 44.65 \\
CCNet \cite{Huang2018CCNetCA} & ResNet-101 & 45.22 \\

\hline
Ours$\dagger$ & ResNet-101 & \textbf{45.30}\\
\midrule
\bottomrule
\end{tabular}
\end{center}

\label{ADE20Kval}
\end{table}

\begin{figure*}[htb]
\centering
\includegraphics[width=1\linewidth]{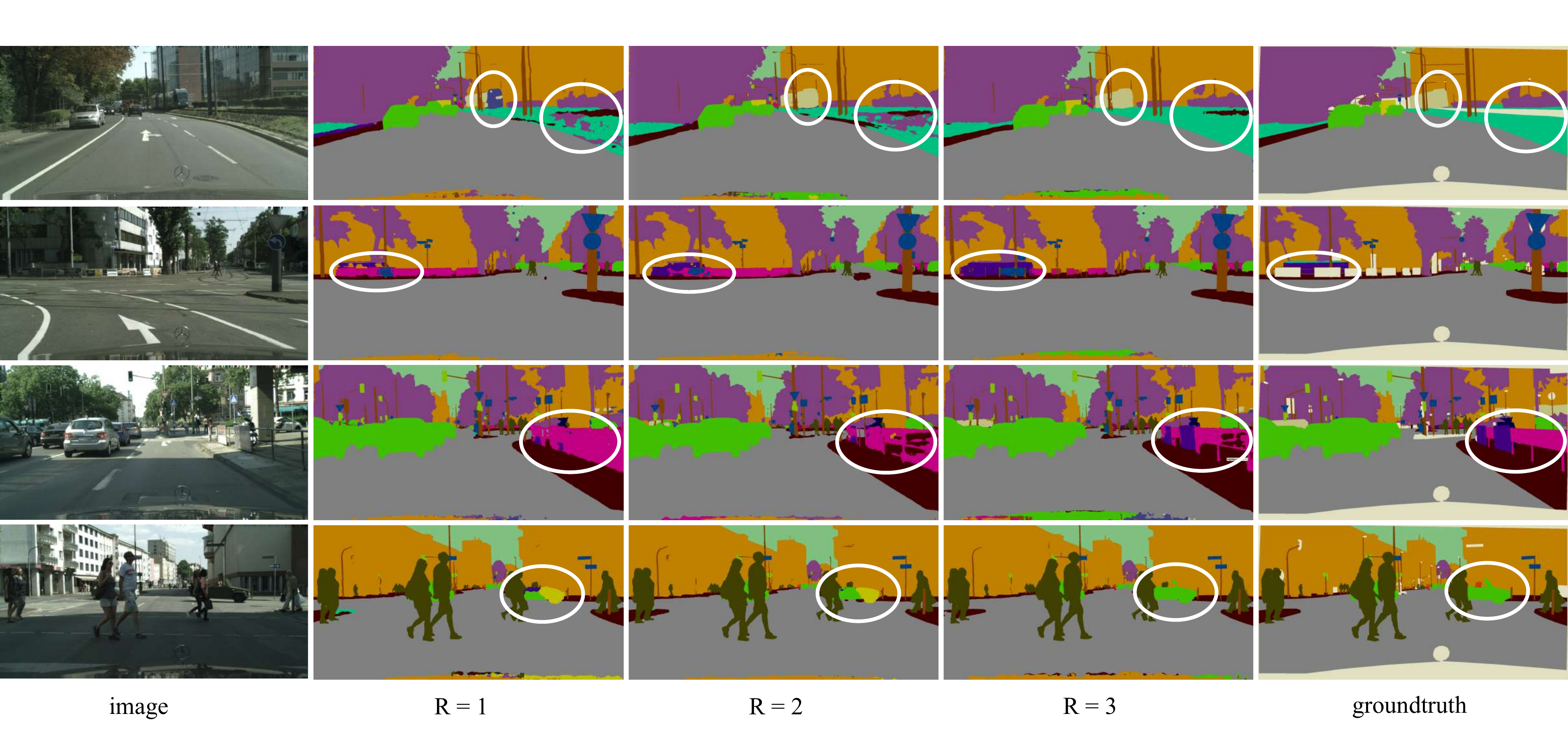}
   \caption{Illustration of the effectiveness of the High-Order Representation module. We leverage the \textit{white circles} to indicate those challenging regions that are easily to be misclassified. The comparison is between different orders in the HR module.}
\label{visualization}
\end{figure*}

\subsubsection{PASCAL Context dataset}
PASCAL Context dataset provides dense semantic labels for the whole scene. We also conduct experiments on the PASCAL Context dataset to verify the generalization of our proposed network. We follow the prior work to use the most frequent 59 object classes plus background as semantic labels. We report the comparison results with the state-of-the-art methods in Table \ref{PASCALval}. With deep pre-trained ResNet-101, our model achieves Mean IoU $52.9\%$, which outperforms previous methods by a large margin. 
\begin{table}[h]
\begin{center}
\caption{Comparison results on PASCAL Context validation set.}
\begin{tabular}{c|c|c}

\toprule
\midrule
Method & Backbone & mIoU(\%) \\
\hline
FCN-8s \cite{Shelhamer2014FullyCN} & - & 37.8\\
Piecewise \cite{lin2016efficient} & - & 43.3\\
DeepLabV2 \cite{chen2017deeplab} & ResNet-101 & 45.7\\
RefineNet \cite{lin2017refinenet} & ResNet-152 & 47.3\\
PSPNet \cite{Zhao2016PyramidSP} & ResNet-101 & 47.8 \\
CCL \cite{ding2018context} & ResNet-101 & 51.6 \\
EncNet \cite{Zhang2018ContextEF} & ResNet-101 & 51.7 \\
DANet \cite{Fu2018DualAN} & ResNet-101 & 52.6 \\

\hline
Ours & ResNet-101 & \textbf{52.9}\\
\midrule
\bottomrule
\end{tabular}
\end{center}

\label{PASCALval}
\end{table}

\subsection{Ablation Study}

In this section, we analyze the High-Order Representation module and High-Order Paired-ASPP module. Ablation experiments are conducted on the validation set of Cityscapes with only single scale testing if not specified. \\
\begin{table}[h]
\begin{center}
\caption{Performance on Cityscapes validation set with various orders.}

\begin{tabular}{c|c|c}

\toprule
\midrule
Orders & Memory(M$\blacktriangle$) & mIoU(\%) \\
\hline
R=1 & 0 & 79.74 \\
R=2 & 58 & 80.15 \\
R=3 & 136 & 80.73 \\
\midrule
\bottomrule
\end{tabular}
\end{center}

\label{highorderablation}
\end{table}

\noindent
\textbf{High-Order Representation module.} In this section we mainly discuss the effect of $R$-th order on network performance. All experiments are conducted with Paired-ASPP module and without multi-scale method. We train our network with various orders in the HR module to find the best implementation. Table \ref{highorderablation} shows the results of our network on Cityscapes validation set by using various orders in the HR module. The input image is $864 \times 864$, resulting in the size of input feature maps $H_i$ of HR module are $432 \times 432$, $216 \times 216$ and $108 \times 108$. The increment of Memory usage is estimated when $R = 1,2,3$.  We can observe that the performance gets better with the increase of the orders. Therefore, we choose $R = 3$ as the default settings in all the following experiments. By increasing the order from $1-th$ to $3-th$ (donated as $R = 1$ and $R = 3$), it improves the performance of $0.99\%$. Results effectively demonstrate the significance of High-Order Representation module. Furthermore, increasing the order $R$ from 1 to 2 can improve the performance by $0.41\%$. So it can be seen that high-order information is useful for contextual capturing. Finally, increasing $R$ from 2 to 3 improves the performance by $0.58\%$. These results prove that the proposed High-Order Representation module can significantly improve the performance by utilizing high-order statistics.

We provide the visualization of results in Figure \ref{visualization} to further illustrate the effectiveness of the High-Order Representation module. We choose some challenging region (the $white circles$ in the images) to illustrate that the confusing regions can be progressively corrected with the order increased. It proves the effectiveness of high-order statistics for semantic segmentation.

\begin{table}[b]
\begin{center}
\caption{Performance on Cityscapes validation set with various combination modes. The combination-1 represents the High-Order Paired-ASPP with ${\mathcal{V}_{3,4}}_{rates=18}, {\mathcal{V}_{2,4}}_{rates=12}$, ${\mathcal{V}_{1,4}}_{rates=6}$. And the combination-2 ${\mathcal{V}_{3,4}}_{rates=6}, {\mathcal{V}_{2,4}}_{rates=12}$, ${\mathcal{V}_{1,4}}_{rates=18}$. Baseline is ResNet-101}
\begin{tabular}{c|c}
\toprule
\midrule
Orders & mIoU(\%) \\
\hline
Baseline  & 75.35\\
combination-1 &  79.74 \\
combination-2  &  79.45\\
\midrule
\bottomrule
\end{tabular}
\end{center}

\label{pairedablation}
\end{table}

\noindent
\textbf{Paired-ASPP module.} In this section we give experiments to verify the effectiveness of the High-Order Paired-ASPP module. The experiments are conducted without High-Order Representation module. Table \ref{pairedablation} illustrates the results of two combination modes in the Paired-ASPP module. In combination-1, we apply atrous convolution with the biggest rate 18 on $\mathcal{V}_{3,4}$ and the smallest rate 6 on $\mathcal{V}_{1,4}$. On the contrary, we apply atrous convolution with the smallest rate 6 on $\mathcal{V}_{3,4}$ and the biggest rate 18 on $\mathcal{V}_{1,4}$ in the combination-2. The combination-1 achieves a good improvement ($75.35\% \rightarrow 79.74\%$) and is better than the combination-2. We suppose that combination-1 can capture much contextual information by providing comprehensive coverage of all scales. 

\begin{figure}[t]
\begin{center}
\includegraphics[width = 8.3cm, height = 4.3cm]{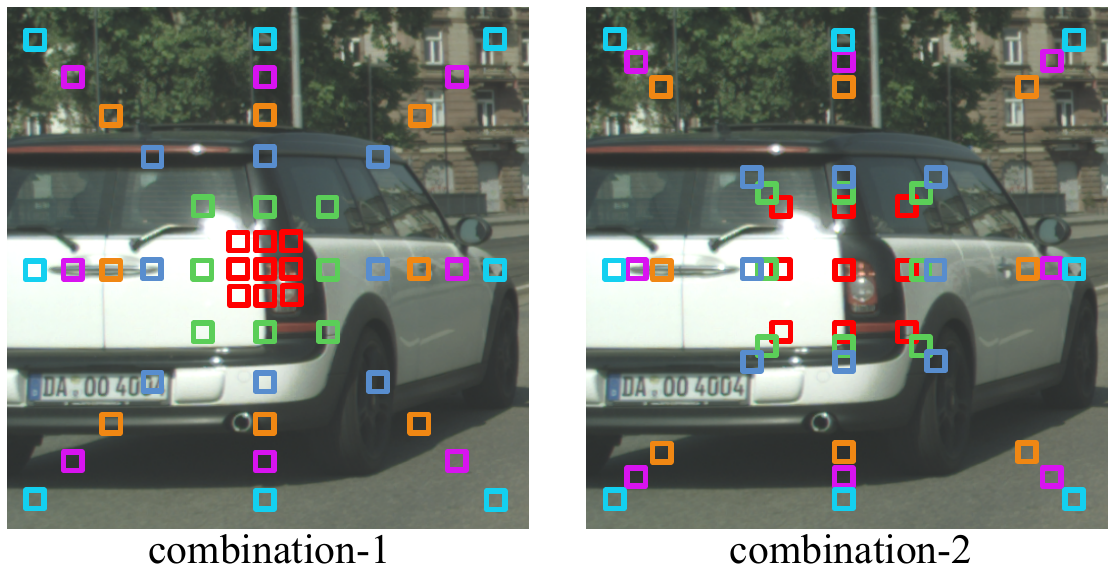}
\end{center}
   \caption{Comparing the scales coverage of combination-1 and combination-2. We can see combination-1 covers more scales than combination-2}
\label{visualizationofpaired}
\end{figure}

In Figure \ref{visualizationofpaired}, we visualize the scales coverage of combination-1 and combination-2. The receptive field becomes larger when the network goes deeper and feature map becomes smaller. Therefore, if we apply atrous convolution with rate 18 on $\mathcal{V}_{1,4}$, we can not obtain the smallest scale information as shown in combination-2 of Figure \ref{visualizationofpaired}. In combination-2, some similar scales information are repeatedly captured. By adopting the combination-1, we can cover the span of scales as large as possible, making features more comprehensive and discriminative.

\noindent
\textbf{Ablation Study for HR and PA modules.} To further verify the performance of the HP and PA modules, we conduct experiments with different settings in Table \ref{Ablation1}.
\begin{table}[ht]
\begin{center}

\caption{Ablation study on Cityscapes val set.}
\begin{tabular}{c c c c | c}

\toprule
\midrule
Backbone & ASPP & PA & HR & mIoU(\%)\\
\hline
ResNet-50 &  & & & 73.24\\
ResNet-50 & $\surd$ & & & 77.91\\
ResNet-50 &  & $\surd$ & & 79.11 \\
ResNet-50 &  & $\surd$ &$\surd$ &  79.88\\
\midrule
ResNet-101  &  & & & 75.35\\
ResNet-101 & $\surd$ & & & 78.90\\
ResNet-101 &  & $\surd$ & & 79.74 \\
ResNet-101 &  & $\surd$ &$\surd$ &  80.73\\

\midrule
\bottomrule
\end{tabular}
\end{center}

\label{Ablation1}
\end{table}
As shown in Table \ref{Ablation1}, the two modules improve the performance remarkably. Compared with the DeepLabV3 (ResNet-50), employing PA module yields a result of $79.11\%$ in mIoU, which brings $1.2\%$ improvement. Meanwhile, employing HR module further improve the performance by $0.77\%$. Moreover, when we adopt the deeper pre-trained ResNet-101, the network with two modules significantly improves the segmentation performance over the DeepLabV3 by $1.83\%$. Results show that PA and HR modules bring great benefit to scene segmentation.

At last, we show the experiments results with different settings in Table \ref{Overall}. We can see that the network with two modules significantly improves the segmentation performance over the DeepLabV3 module by $1.83\%$. The results demonstrate the efficacy of the proposed High-Order Representation module and High-Order Paired-ASPP module.

\begin{table}[ht]
\begin{center}

\caption{Performance comparison between different strategies on Cityscapes val set. HP represents High-Order Paired-ASPP Network, MS represents multi-scale method and Flip represents horizontal flip method.}
\begin{tabular}{c c c c c | c}
\toprule
\midrule
Method & ASPP & HP & MS & Flip & mIoU(\%)\\
\hline
Baseline  & & & & & 75.35 \\ 
DeepLabV3 \cite{Chen2017RethinkingAC} & $\surd$ & & & & 78.90\\
HPNet  &  &$\surd$ & & & 80.73 \\
HPNet  & & $\surd$ & $\surd$ & &  81.60\\
HPNet  & &  $\surd$ & $\surd$ & $\surd$ & 81.88 \\

\midrule
\bottomrule
\end{tabular}
\end{center}

\label{Overall}
\end{table}

\section{Conclusion}
In this paper, we propose the High-Order Paired-ASPP networks, which is a new CNN architecture for semantic segmentation by exploiting high-order statistics and enhancing the feature fusion. This approach can generate low-dimensional high-order statistics from different layers. The high-order representation is driven by convolutional activations, via high-order polynomial predictor, we capture discriminative features. Unlike existing methods, we make full use of the low-level features and extract high-order spatial information for later feature fusion. In Paired-ASPP module, we fuse the high-order information with a novel paired manner to get multi-scale information. The Paired-ASPP can capture more scales information than normal ASPP module, thus showing higher mIoU than DeepLabV3 and other state-of-the-art methods. 

To evaluate the proposed method, this study conducted experiments on three benchmarks, Cityscapes, ADE20K and Pascal-Context, in which the highest mIoU is shown when using 3-th order and ResNet-101 as the backbone. Particularly, we've already conducted the experiments of 4-th order in HR module. Compare to 3-th order, it brings only small improvement but huge computation cost. To balance the performance and resource usage, we choose 3-th order as our default settings. Our experiments show that high-order statistics contain more discriminative information and can correct the confusing regions. Our network achieves state-of-the-art results on three challenging datasets (81.6\% with Cityscapes, 45.30\% with ADE20K and 52.9\% with Pascal-Context.)

Our work suggests promising future directions of exploiting relationships between low-level features and high-order statistics to advance the state-of-the-art in segmentation tasks. 

\ifCLASSOPTIONcaptionsoff
  \newpage
\fi



%


%


\begin{thebibliography}{10}
\providecommand{\url}[1]{#1}
\csname url@samestyle\endcsname
\providecommand{\newblock}{\relax}
\providecommand{\bibinfo}[2]{#2}
\providecommand{\BIBentrySTDinterwordspacing}{\spaceskip=0pt\relax}
\providecommand{\BIBentryALTinterwordstretchfactor}{4}
\providecommand{\BIBentryALTinterwordspacing}{\spaceskip=\fontdimen2\font plus
\BIBentryALTinterwordstretchfactor\fontdimen3\font minus
  \fontdimen4\font\relax}
\providecommand{\BIBforeignlanguage}[2]{{%
\expandafter\ifx\csname l@#1\endcsname\relax
\typeout{** WARNING: IEEEtran.bst: No hyphenation pattern has been}%
\typeout{** loaded for the language `#1'. Using the pattern for}%
\typeout{** the default language instead.}%
\else
\language=\csname l@#1\endcsname
\fi
#2}}
\providecommand{\BIBdecl}{\relax}
\BIBdecl

\bibitem{Shelhamer2014FullyCN}
E.~Shelhamer, J.~Long, and T.~Darrell, ``Fully convolutional networks for
  semantic segmentation,'' \emph{IEEE Transactions on Pattern Analysis and
  Machine Intelligence}, vol.~39, pp. 640--651, 2014.

\bibitem{Xu2016EndtoEndLO}
H.~Xu, Y.~Gao, F.~Yu, and T.~Darrell, ``End-to-end learning of driving models
  from large-scale video datasets,'' \emph{2017 IEEE Conference on Computer
  Vision and Pattern Recognition (CVPR)}, pp. 3530--3538, 2016.

\bibitem{Litjens2017ASO}
G.~J.~S. Litjens, T.~Kooi, B.~E. Bejnordi, A.~A.~A. Setio, F.~Ciompi,
  M.~Ghafoorian, J.~van~der Laak, B.~van Ginneken, and C.~I. S{\'a}nchez, ``A
  survey on deep learning in medical image analysis,'' \emph{Medical image
  analysis}, vol.~42, pp. 60--88, 2017.

\bibitem{kampffmeyer2016semantic}
M.~Kampffmeyer, A.-B. Salberg, and R.~Jenssen, ``Semantic segmentation of small
  objects and modeling of uncertainty in urban remote sensing images using deep
  convolutional neural networks,'' in \emph{Proceedings of the IEEE conference
  on computer vision and pattern recognition workshops}, 2016, pp. 1--9.

\bibitem{Chen2017RethinkingAC}
L.-C. Chen, G.~Papandreou, F.~Schroff, and H.~Adam, ``Rethinking atrous
  convolution for semantic image segmentation,'' \emph{ArXiv}, vol.
  abs/1706.05587, 2017.

\bibitem{Zhao2016PyramidSP}
H.~Zhao, J.~Shi, X.~Qi, X.~Wang, and J.~Jia, ``Pyramid scene parsing network,''
  \emph{2017 IEEE Conference on Computer Vision and Pattern Recognition
  (CVPR)}, pp. 6230--6239, 2016.

\bibitem{Zhang2018ContextEF}
H.~Zhang, K.~J. Dana, J.~Shi, Z.~Zhang, X.~Wang, A.~Tyagi, and A.~Agrawal,
  ``Context encoding for semantic segmentation,'' \emph{2018 IEEE/CVF
  Conference on Computer Vision and Pattern Recognition}, pp. 7151--7160, 2018.

\bibitem{Vaswani2017AttentionIA}
A.~Vaswani, N.~Shazeer, N.~Parmar, J.~Uszkoreit, L.~Jones, A.~N. Gomez,
  L.~Kaiser, and I.~Polosukhin, ``Attention is all you need,'' in \emph{NIPS},
  2017.

\bibitem{Fu2018DualAN}
J.~Fu, J.~Liu, H.~Tian, Z.~Fang, and H.~Lu, ``Dual attention network for scene
  segmentation,'' \emph{ArXiv}, vol. abs/1809.02983, 2018.

\bibitem{Huang2018CCNetCA}
Z.~Huang, X.~Wang, L.~Huang, C.~Huang, Y.~Wei, and W.~Liu, ``Ccnet: Criss-cross
  attention for semantic segmentation,'' \emph{ArXiv}, vol. abs/1811.11721,
  2018.

\bibitem{Li2019ExpectationMaximizationAN}
X.~Li, Z.~Zhong, J.~Wu, Y.~Yang, Z.~Lin, and H.~Liu, ``Expectation-maximization
  attention networks for semantic segmentation,'' \emph{ArXiv}, vol.
  abs/1907.13426, 2019.

\bibitem{zhang2018exfuse}
Z.~Zhang, X.~Zhang, C.~Peng, X.~Xue, and J.~Sun, ``Exfuse: Enhancing feature
  fusion for semantic segmentation,'' in \emph{Proceedings of the European
  Conference on Computer Vision (ECCV)}, 2018, pp. 269--284.

\bibitem{li2019dfanet}
H.~Li, P.~Xiong, H.~Fan, and J.~Sun, ``Dfanet: Deep feature aggregation for
  real-time semantic segmentation,'' in \emph{Proceedings of the IEEE
  Conference on Computer Vision and Pattern Recognition}, 2019, pp. 9522--9531.

\bibitem{takikawa2019gated}
T.~Takikawa, D.~Acuna, V.~Jampani, and S.~Fidler, ``Gated-scnn: Gated shape
  cnns for semantic segmentation,'' \emph{arXiv preprint arXiv:1907.05740},
  2019.

\bibitem{yu2018learning}
C.~Yu, J.~Wang, C.~Peng, C.~Gao, G.~Yu, and N.~Sang, ``Learning a
  discriminative feature network for semantic segmentation,'' in
  \emph{Proceedings of the IEEE Conference on Computer Vision and Pattern
  Recognition}, 2018, pp. 1857--1866.

\bibitem{Cai2017HigherOrderIO}
S.~Cai, W.~Zuo, and L.~Zhang, ``Higher-order integration of hierarchical
  convolutional activations for fine-grained visual categorization,''
  \emph{2017 IEEE International Conference on Computer Vision (ICCV)}, pp.
  511--520, 2017.

\bibitem{badrinarayanan2017segnet}
V.~Badrinarayanan, A.~Kendall, and R.~Cipolla, ``Segnet: A deep convolutional
  encoder-decoder architecture for image segmentation,'' \emph{IEEE
  transactions on pattern analysis and machine intelligence}, vol.~39, no.~12,
  pp. 2481--2495, 2017.

\bibitem{noh2015learning}
H.~Noh, S.~Hong, and B.~Han, ``Learning deconvolution network for semantic
  segmentation,'' in \emph{Proceedings of the IEEE international conference on
  computer vision}, 2015, pp. 1520--1528.

\bibitem{ronneberger2015u}
O.~Ronneberger, P.~Fischer, and T.~Brox, ``U-net: Convolutional networks for
  biomedical image segmentation,'' in \emph{International Conference on Medical
  image computing and computer-assisted intervention}.\hskip 1em plus 0.5em
  minus 0.4em\relax Springer, 2015, pp. 234--241.

\bibitem{ghiasi2016laplacian}
G.~Ghiasi and C.~C. Fowlkes, ``Laplacian pyramid reconstruction and refinement
  for semantic segmentation,'' in \emph{European Conference on Computer
  Vision}.\hskip 1em plus 0.5em minus 0.4em\relax Springer, 2016, pp. 519--534.

\bibitem{amirul2017gated}
M.~Amirul~Islam, M.~Rochan, N.~D. Bruce, and Y.~Wang, ``Gated feedback
  refinement network for dense image labeling,'' in \emph{Proceedings of the
  IEEE Conference on Computer Vision and Pattern Recognition}, 2017, pp.
  3751--3759.

\bibitem{lin2017refinenet}
G.~Lin, A.~Milan, C.~Shen, and I.~Reid, ``Refinenet: Multi-path refinement
  networks for high-resolution semantic segmentation,'' in \emph{Proceedings of
  the IEEE conference on computer vision and pattern recognition}, 2017, pp.
  1925--1934.

\bibitem{zhao2018icnet}
H.~Zhao, X.~Qi, X.~Shen, J.~Shi, and J.~Jia, ``Icnet for real-time semantic
  segmentation on high-resolution images,'' in \emph{Proceedings of the
  European Conference on Computer Vision (ECCV)}, 2018, pp. 405--420.

\bibitem{jegou2017one}
S.~J{\'e}gou, M.~Drozdzal, D.~Vazquez, A.~Romero, and Y.~Bengio, ``The one
  hundred layers tiramisu: Fully convolutional densenets for semantic
  segmentation,'' in \emph{Proceedings of the IEEE conference on computer
  vision and pattern recognition workshops}, 2017, pp. 11--19.

\bibitem{Huang2016DenselyCC}
G.~Huang, Z.~Liu, and K.~Q. Weinberger, ``Densely connected convolutional
  networks,'' \emph{2017 IEEE Conference on Computer Vision and Pattern
  Recognition (CVPR)}, pp. 2261--2269, 2016.

\bibitem{lafferty2001conditional}
J.~Lafferty, A.~McCallum, and F.~C. Pereira, ``Conditional random fields:
  Probabilistic models for segmenting and labeling sequence data,'' 2001.

\bibitem{chandra2016fast}
S.~Chandra and I.~Kokkinos, ``Fast, exact and multi-scale inference for
  semantic image segmentation with deep gaussian crfs,'' in \emph{European
  Conference on Computer Vision}.\hskip 1em plus 0.5em minus 0.4em\relax
  Springer, 2016, pp. 402--418.

\bibitem{Chen2014SemanticIS}
L.-C. Chen, G.~Papandreou, I.~Kokkinos, K.~Murphy, and A.~L. Yuille, ``Semantic
  image segmentation with deep convolutional nets and fully connected crfs,''
  \emph{CoRR}, vol. abs/1412.7062, 2014.

\bibitem{he2015spatial}
K.~He, X.~Zhang, S.~Ren, and J.~Sun, ``Spatial pyramid pooling in deep
  convolutional networks for visual recognition,'' \emph{IEEE transactions on
  pattern analysis and machine intelligence}, vol.~37, no.~9, pp. 1904--1916,
  2015.

\bibitem{chen2017deeplab}
L.-C. Chen, G.~Papandreou, I.~Kokkinos, K.~Murphy, and A.~L. Yuille, ``Deeplab:
  Semantic image segmentation with deep convolutional nets, atrous convolution,
  and fully connected crfs,'' \emph{IEEE transactions on pattern analysis and
  machine intelligence}, vol.~40, no.~4, pp. 834--848, 2017.

\bibitem{vemulapalli2016gaussian}
R.~Vemulapalli, O.~Tuzel, M.-Y. Liu, and R.~Chellapa, ``Gaussian conditional
  random field network for semantic segmentation,'' in \emph{Proceedings of the
  IEEE conference on computer vision and pattern recognition}, 2016, pp.
  3224--3233.

\bibitem{Peng2017LargeKM}
C.~Peng, X.~Zhang, G.~Yu, G.~Luo, and J.~Sun, ``Large kernel matters —
  improve semantic segmentation by global convolutional network,'' \emph{2017
  IEEE Conference on Computer Vision and Pattern Recognition (CVPR)}, pp.
  1743--1751, 2017.

\bibitem{liu2015parsenet}
W.~Liu, A.~Rabinovich, and A.~C. Berg, ``Parsenet: Looking wider to see
  better,'' \emph{arXiv preprint arXiv:1506.04579}, 2015.

\bibitem{shen2018disan}
T.~Shen, T.~Zhou, G.~Long, J.~Jiang, S.~Pan, and C.~Zhang, ``Disan: Directional
  self-attention network for rnn/cnn-free language understanding,'' in
  \emph{Thirty-Second AAAI Conference on Artificial Intelligence}, 2018.

\bibitem{lin2017structured}
Z.~Lin, M.~Feng, C.~N.~d. Santos, M.~Yu, B.~Xiang, B.~Zhou, and Y.~Bengio, ``A
  structured self-attentive sentence embedding,'' \emph{arXiv preprint
  arXiv:1703.03130}, 2017.

\bibitem{wang2018non}
X.~Wang, R.~Girshick, A.~Gupta, and K.~He, ``Non-local neural networks,'' in
  \emph{Proceedings of the IEEE conference on computer vision and pattern
  recognition}, 2018, pp. 7794--7803.

\bibitem{zhu2019asymmetric}
Z.~Zhu, M.~Xu, S.~Bai, T.~Huang, and X.~Bai, ``Asymmetric non-local neural
  networks for semantic segmentation,'' \emph{arXiv preprint arXiv:1908.07678},
  2019.

\bibitem{yang2018denseaspp}
M.~Yang, K.~Yu, C.~Zhang, Z.~Li, and K.~Yang, ``Denseaspp for semantic
  segmentation in street scenes,'' in \emph{Proceedings of the IEEE Conference
  on Computer Vision and Pattern Recognition}, 2018, pp. 3684--3692.

\bibitem{cai2017higher}
S.~Cai, W.~Zuo, and L.~Zhang, ``Higher-order integration of hierarchical
  convolutional activations for fine-grained visual categorization,'' in
  \emph{Proceedings of the IEEE International Conference on Computer Vision},
  2017, pp. 511--520.

\bibitem{cui2017kernel}
Y.~Cui, F.~Zhou, J.~Wang, X.~Liu, Y.~Lin, and S.~Belongie, ``Kernel pooling for
  convolutional neural networks,'' in \emph{Proceedings of the IEEE conference
  on computer vision and pattern recognition}, 2017, pp. 2921--2930.

\bibitem{ionescu2015matrix}
C.~Ionescu, O.~Vantzos, and C.~Sminchisescu, ``Matrix backpropagation for deep
  networks with structured layers,'' in \emph{Proceedings of the IEEE
  International Conference on Computer Vision}, 2015, pp. 2965--2973.

\bibitem{lin2015bilinear}
T.-Y. Lin, A.~RoyChowdhury, and S.~Maji, ``Bilinear cnn models for fine-grained
  visual recognition,'' in \emph{Proceedings of the IEEE international
  conference on computer vision}, 2015, pp. 1449--1457.

\bibitem{li2017second}
P.~Li, J.~Xie, Q.~Wang, and W.~Zuo, ``Is second-order information helpful for
  large-scale visual recognition?'' in \emph{Proceedings of the IEEE
  International Conference on Computer Vision}, 2017, pp. 2070--2078.

\bibitem{Wang2017G2DeNetGG}
Q.~Wang, P.~Li, and L.~Zhang, ``G2denet: Global gaussian distribution embedding
  network and its application to visual recognition,'' \emph{2017 IEEE
  Conference on Computer Vision and Pattern Recognition (CVPR)}, pp.
  6507--6516, 2017.

\bibitem{wang2018multi}
H.~Wang, Q.~Wang, M.~Gao, P.~Li, and W.~Zuo, ``Multi-scale location-aware
  kernel representation for object detection,'' in \emph{Proceedings of the
  IEEE Conference on Computer Vision and Pattern Recognition}, 2018, pp.
  1248--1257.

\bibitem{chen2019mixed}
B.~Chen, W.~Deng, and J.~Hu, ``Mixed high-order attention network for person
  re-identification,'' \emph{arXiv preprint arXiv:1908.05819}, 2019.

\bibitem{kolda2009tensor}
T.~G. Kolda and B.~W. Bader, ``Tensor decompositions and applications,''
  \emph{SIAM review}, vol.~51, no.~3, pp. 455--500, 2009.

\bibitem{he2016deep}
K.~He, X.~Zhang, S.~Ren, and J.~Sun, ``Deep residual learning for image
  recognition,'' in \emph{Proceedings of the IEEE conference on computer vision
  and pattern recognition}, 2016, pp. 770--778.

\bibitem{cordts2016cityscapes}
M.~Cordts, M.~Omran, S.~Ramos, T.~Rehfeld, M.~Enzweiler, R.~Benenson,
  U.~Franke, S.~Roth, and B.~Schiele, ``The cityscapes dataset for semantic
  urban scene understanding,'' in \emph{Proceedings of the IEEE conference on
  computer vision and pattern recognition}, 2016, pp. 3213--3223.

\bibitem{zhou2017scene}
B.~Zhou, H.~Zhao, X.~Puig, S.~Fidler, A.~Barriuso, and A.~Torralba, ``Scene
  parsing through ade20k dataset,'' in \emph{Proceedings of the IEEE conference
  on computer vision and pattern recognition}, 2017, pp. 633--641.

\bibitem{mottaghi2014role}
R.~Mottaghi, X.~Chen, X.~Liu, N.-G. Cho, S.-W. Lee, S.~Fidler, R.~Urtasun, and
  A.~Yuille, ``The role of context for object detection and semantic
  segmentation in the wild,'' in \emph{Proceedings of the IEEE Conference on
  Computer Vision and Pattern Recognition}, 2014, pp. 891--898.

\bibitem{chen2018encoder}
L.-C. Chen, Y.~Zhu, G.~Papandreou, F.~Schroff, and H.~Adam, ``Encoder-decoder
  with atrous separable convolution for semantic image segmentation,'' in
  \emph{Proceedings of the European conference on computer vision (ECCV)},
  2018, pp. 801--818.

\bibitem{chen2018searching}
L.-C. Chen, M.~Collins, Y.~Zhu, G.~Papandreou, B.~Zoph, F.~Schroff, H.~Adam,
  and J.~Shlens, ``Searching for efficient multi-scale architectures for dense
  image prediction,'' in \emph{Advances in Neural Information Processing
  Systems}, 2018, pp. 8699--8710.

\bibitem{wu2019wider}
Z.~Wu, C.~Shen, and A.~Van Den~Hengel, ``Wider or deeper: Revisiting the resnet
  model for visual recognition,'' \emph{Pattern Recognition}, vol.~90, pp.
  119--133, 2019.

\bibitem{wang2018understanding}
P.~Wang, P.~Chen, Y.~Yuan, D.~Liu, Z.~Huang, X.~Hou, and G.~Cottrell,
  ``Understanding convolution for semantic segmentation,'' in \emph{2018 IEEE
  winter conference on applications of computer vision (WACV)}.\hskip 1em plus
  0.5em minus 0.4em\relax IEEE, 2018, pp. 1451--1460.

\bibitem{zhang2017scale}
R.~Zhang, S.~Tang, Y.~Zhang, J.~Li, and S.~Yan, ``Scale-adaptive convolutions
  for scene parsing,'' in \emph{Proceedings of the IEEE International
  Conference on Computer Vision}, 2017, pp. 2031--2039.

\bibitem{yu2018bisenet}
C.~Yu, J.~Wang, C.~Peng, C.~Gao, G.~Yu, and N.~Sang, ``Bisenet: Bilateral
  segmentation network for real-time semantic segmentation,'' in
  \emph{Proceedings of the European Conference on Computer Vision (ECCV)},
  2018, pp. 325--341.

\bibitem{ke2018adaptive}
T.-W. Ke, J.-J. Hwang, Z.~Liu, and S.~X. Yu, ``Adaptive affinity fields for
  semantic segmentation,'' in \emph{Proceedings of the European Conference on
  Computer Vision (ECCV)}, 2018, pp. 587--602.

\bibitem{zhao2018psanet}
H.~Zhao, Y.~Zhang, S.~Liu, J.~Shi, C.~Change~Loy, D.~Lin, and J.~Jia, ``Psanet:
  Point-wise spatial attention network for scene parsing,'' in
  \emph{Proceedings of the European Conference on Computer Vision (ECCV)},
  2018, pp. 267--283.

\bibitem{xiao2018unified}
T.~Xiao, Y.~Liu, B.~Zhou, Y.~Jiang, and J.~Sun, ``Unified perceptual parsing
  for scene understanding,'' in \emph{Proceedings of the European Conference on
  Computer Vision (ECCV)}, 2018, pp. 418--434.

\bibitem{liang2018dynamic}
X.~Liang, H.~Zhou, and E.~Xing, ``Dynamic-structured semantic propagation
  network,'' in \emph{Proceedings of the IEEE Conference on Computer Vision and
  Pattern Recognition}, 2018, pp. 752--761.

\bibitem{lin2016efficient}
G.~Lin, C.~Shen, A.~Van Den~Hengel, and I.~Reid, ``Efficient piecewise training
  of deep structured models for semantic segmentation,'' in \emph{Proceedings
  of the IEEE conference on computer vision and pattern recognition}, 2016, pp.
  3194--3203.

\bibitem{ding2018context}
H.~Ding, X.~Jiang, B.~Shuai, A.~Qun~Liu, and G.~Wang, ``Context contrasted
  feature and gated multi-scale aggregation for scene segmentation,'' in
  \emph{Proceedings of the IEEE Conference on Computer Vision and Pattern
  Recognition}, 2018, pp. 2393--2402.

\end{thebibliography}

\bibliographystyle{IEEEtran}

%








\end{document}